\pdfoutput=1

\documentclass[11pt]{article}

\usepackage{ACL2023}

\usepackage{times}
\usepackage{latexsym}
\usepackage{bm}
\usepackage{algorithmic}
\usepackage{algorithm}
\usepackage{graphicx}
\usepackage{tabularx}
\usepackage{amsmath}
\usepackage{booktabs}

\usepackage[T1]{fontenc}

\usepackage[utf8]{inputenc}

\usepackage{microtype}

\usepackage{inconsolata}

\DeclareMathOperator*{\argmax}{argmax}
\DeclareMathOperator*{\argmin}{argmin}

\newcolumntype{C}{>{\centering\arraybackslash}X}
\newcolumntype{L}{>{\raggedright\arraybackslash}X}
\newcolumntype{R}{>{\raggedleft\arraybackslash}X}

\newcommand{\substitute}[2]{\left.#1\right|_{#2}}

\title{Differentiable Instruction Optimization for Cross-Task Generalization}

\author{
Masaru Isonuma$^{1,2}$ \quad 
Junichiro Mori$^{1,3}$ \quad 
Ichiro Sakata$^{1}$ \bigskip\\
$^1$ The University of Tokyo \quad 
$^2$ The University of Edinburgh \quad 
$^3$ RIKEN \\
  {\tt \{isonuma, isakata\}@ipr-ctr.t.u-tokyo.ac.jp \quad mori@mi.u-tokyo.ac.jp}
}
 
\begin{document}
\maketitle
\begin{abstract}
Instruction tuning has been attracting much attention to achieve generalization ability across a wide variety of tasks.
Although various types of instructions have been manually created for instruction tuning, it is still unclear what kind of instruction is optimal to obtain cross-task generalization ability.
This work presents \emph{instruction optimization}, which optimizes training instructions with respect to generalization ability.
Rather than manually tuning instructions, we introduce learnable instructions and optimize them with gradient descent by leveraging bilevel optimization.
Experimental results show that the learned instruction enhances the diversity of instructions and improves the generalization ability compared to using only manually created instructions.

\end{abstract}

\section{Introduction}

Recently, significant progress has been made in developing models that can generalize to arbitrary tasks by following natural language descriptions \cite{brown2020language, ouyang2022training}. 
\emph{Instruction tuning} has been a region of interest as a training technique to obtain such generalization ability \cite{wei2021finetuned,sanh2021multitask,mishra-etal-2022-cross}.
By finetuning pretrained language models on a variety of tasks with their instructions, models can generalize to arbitrary tasks unseen during training.
Many previous studies witnessed the effectiveness of instruction tuning \cite{chung2022scaling, wang-etal-2022-super, lampinen-etal-2022-language}.

Various instructions have been created for instruction tuning, such as task name, task definition, positive/negative exemplars of a task, explanations of why each positive/negative exemplar is correct/incorrect, etc.
However, \citet{mishra-etal-2022-cross, wang-etal-2022-super} showed that the definition and positive exemplars of tasks are sufficient for instruction tuning, and the effect of adding other types of instruction is negligible or sometimes has a negative impact on the generalization performance.
Seeking an optimal instruction for cross-task generalization is an important issue for instruction tuning, while it requires much human effort (100+ researchers have participated in previous studies).
Furthermore, human-interpretable instructions are not necessarily optimal for obtaining cross-task generalization ability.

\begin{figure}[t!]
\centering
\includegraphics[width=\linewidth]{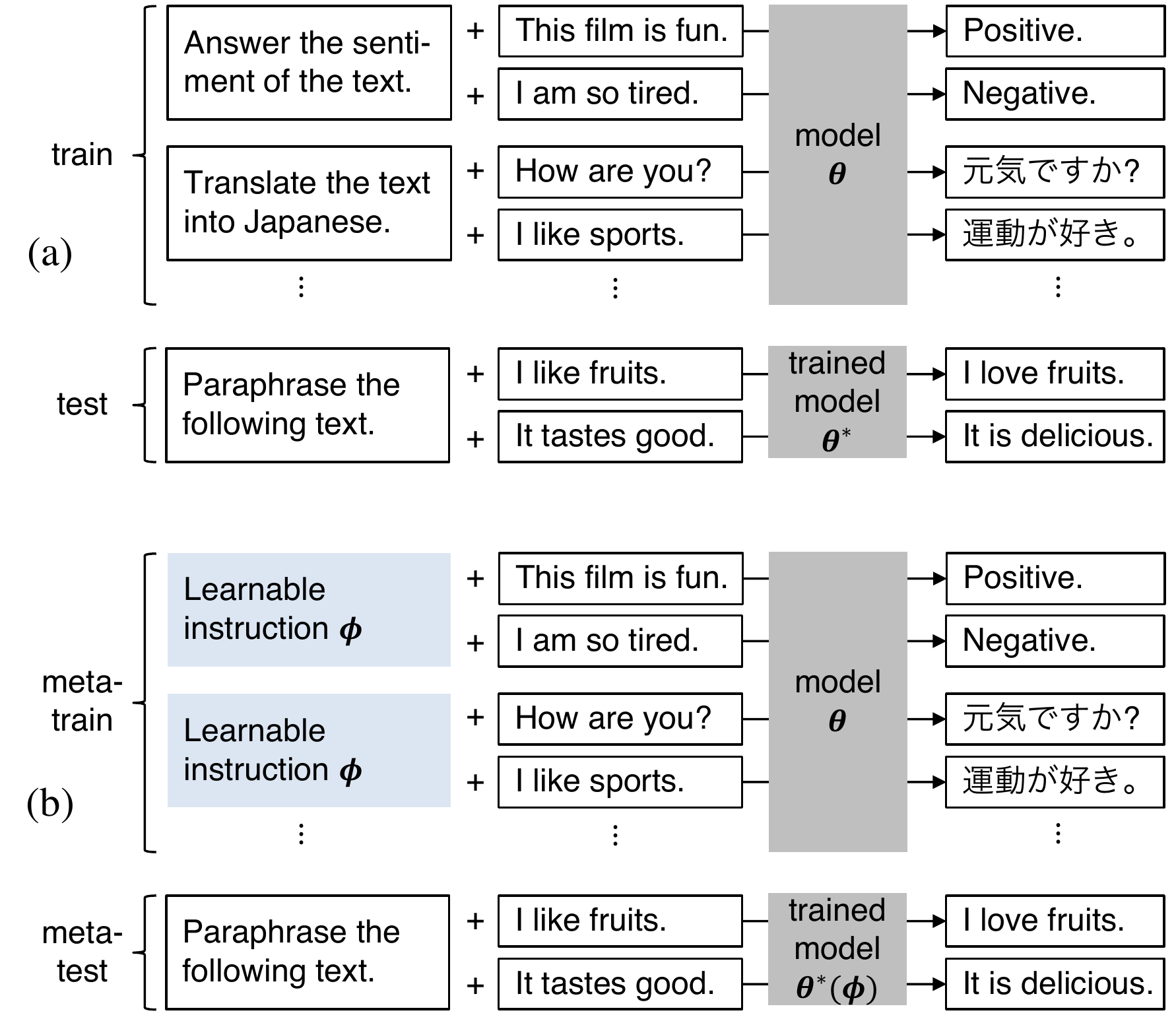}
\caption{Outline of (a) instruction tuning and (b) instruction optimization (ours).}
\label{fig:introduction}
\end{figure}

Against this background, we propose \emph{instruction optimization}, which introduces learnable instructions and optimizes them w.r.t. the cross-task generalization ability.
As shown in Figure \ref{fig:introduction}, a model $\bm{\theta}$ is optimized to maximize the performance on meta-train tasks following learnable instructions.
By contrast, learnable instructions $\bm{\phi}$ are trained to maximize the meta-test performance of the trained model $\bm{\theta}^{*}(\bm{\phi})$.
This optimization is called bilevel optimization and is frequently used in hyperparameter optimization \cite{franceschi2017forward, lorraine2020optimizing}, meta-learning \cite{finn2017model, franceschi2018bilevel}, and neural architecture search \cite{liu2018darts, zhang2021idarts}.
We regard training instructions as a special type of hyperparameter and optimize them with gradient descent by relaxing the search space to be continuous.

To create learnable instructions, we propose two methods: \emph{instruction embedder}, which generates the embeddings of instructions, and \emph{instruction extractor}, which selects an optimal task exemplar.
Recently, prompt engineering has drawn attention to seek the optimal prompt to achieve a task \cite{liu2022pretrain}.
Some work studies continuous prompts that perform prompting in the embedding space of tokens \cite{li-liang-2021-prefix, lester-etal-2021-power}, whereas others retrieve optimal exemplars as a testing prompt for in-context learning \cite{liu-etal-2022-makes, rubin-etal-2022-learning}.
Our instruction embedder and instruction extractor follow the idea of continuous prompts and prompt retrievers, respectively.
Whereas previous work optimizes prompts to solve an individual task on the test, our study differs in the target and aim of optimization.
We optimize the training prompts to maximize the cross-task generalization ability of the trained model.

In the experiment, we confirmed that the instruction extractor successfully extracted appropriate instruction, providing proof of concept.
Regarding the comparison with instruction tuning, the instruction embedder enhances the diversity of instructions and improves the generalization ability compared to using only manually created instructions.
In contrast, the instruction extractor does not contributes to the performance gain, which shows that using the same task exemplar across instances is unexpectedly preferable for cross-task generalization.
This study provides a basis for exploring the optimal instructions for instruction tuning.

\begin{figure*}[t!]
\centering
\includegraphics[width=\linewidth]{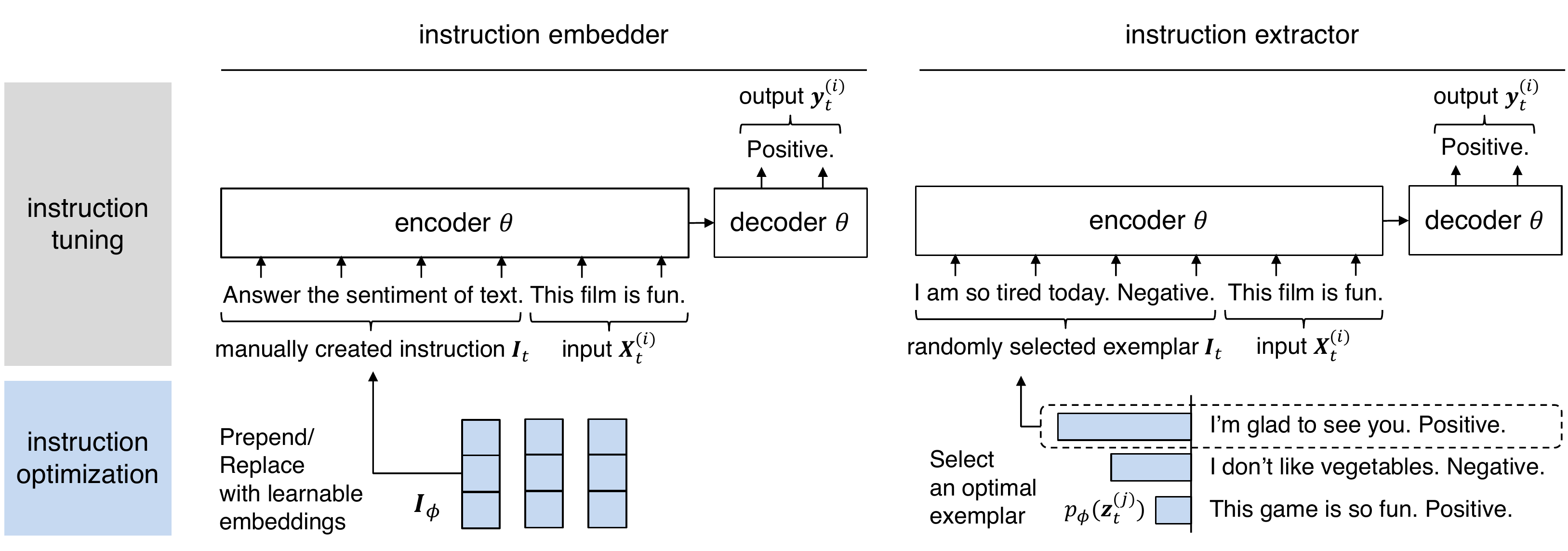}
\caption{Outline of instruction embedder and instruction extractor. Instruction tuning uses a manually created instruction or randomly selected exemplar as \emph{training} instruction. In contrast, instruction embedder introduces the learnable embeddings of instruction, while instruction extractor selects an optimal exemplar as \emph{training} instruction.}
\label{fig:instructionoptimization}
\end{figure*}

\section{Preliminaries}

Instruction tuning trains a model $\bm{\theta}$ to minimize the training loss defined in Eq. \eqref{eq:train}:
\begin{align}
\begin{split}
\label{eq:train}
\bm{\theta}^{*} 
&\!=\! \argmin_{\bm{\theta}} \mathcal{L}(\bm{\theta}) \\
&\!=\! \argmin_{\bm{\theta}} \sum_{t \in \mathcal{T}_{train}} \sum_{i=1}^{N_t} \!-\!\log p_{\bm{\theta}}(\bm{y}_t^{(i)}|[\bm{I}_t; \bm{X}_t^{(i)}]) 
\end{split}
\end{align}
where $\bm{X}_t^{(i)}$ and $\bm{I}_t$ denote the embedding matrix of the $i$-th input and instruction of the task $t$, respectively.
$\bm{y}_t^{(i)}$ is a sequence of tokens that represents a class label or reference text.
Instruction tuning regards all tasks as the conditional text generation given the concatenation of the instruction and task input $[\bm{I}_t; \bm{X}_t]$.
By prepending the instruction to the task input, the trained model $\bm{\theta}^{*}$ can generalize to a variety of unseen tasks $t \notin \mathcal{T}_{train}$.

The optimal training instructions have been sought by manually creating various types of instruction for instruction tuning \cite{mishra-etal-2022-cross, wei2021finetuned, sanh2021multitask}.
However, \citet{mishra-etal-2022-cross, wang-etal-2022-super} showed that task definition and task exemplars are sufficient for instruction tuning, while adding other types of instruction is negligible or sometimes negatively affects the generalization performance.
This observation motivates us to automatically optimize training instructions, rather than manually tuning them.
We introduce learnable instructions and optimize them with gradient descent by leveraging bilevel optimization.
The next section provides the details of instruction optimization.

\section{Instruction Optimization}

Instruction optimization splits training tasks $\mathcal{T}_{train}$ into two sets: meta-train tasks $\mathcal{T}_{meta-train}$ and meta-test tasks $\mathcal{T}_{meta-test}$.
Subsequently, a model $\bm{\theta}$ is trained to minimize the inner loss on meta-train tasks following learnable instructions $\bm{I}_{\phi}$ in Eq. \eqref{eq:inner}.
\begin{align}
\begin{split}
&\bm{\theta}^{*}(\bm{\phi}) = \argmin_{\bm{\theta}} \mathcal{L}_{in}(\bm{\theta}, \bm{\phi}) \\
&\!=\! \argmin_{\bm{\theta}} \! \sum_{t \in \mathcal{T}_{meta-train}} \sum_{i=1}^{N_t} \!-\! \log p_{\bm{\theta}}(\bm{y}_t^{(i)}|[\bm{I}_\phi; \bm{X}_t^{(i)}])
\end{split}
\label{eq:inner}
\end{align}
where $\bm{\phi}$ is a parameter for learnable instructions.
$\bm{I}_\phi$ is constructed using an instruction embedder (Section \ref{sec:instructionembedder}) or an instruction extractor (Section \ref{sec:instructionextractor}), which will be explained later.

If the learnable instruction $\bm{I}_\phi$ is randomly created, the trained model $\bm{\theta}^{*}(\bm{\phi})$ performs poorly on unseen tasks. 
Therefore, we optimize $\bm{\phi}$ such that the trained model $\bm{\theta}^{*}(\bm{\phi})$ achieves high performance on meta-test tasks, which are not shown during training.
$\bm{\phi}$ is updated to minimize the outer loss in Eq. \eqref{eq:outer}.
\begin{align}
\begin{split}
&\bm{\phi}^{*}
= \argmin_{\bm{\phi}} \mathcal{L}_{out}(\bm{\theta}^{*}(\bm{\phi})) \\
&\!=\! \argmin_{\bm{\phi}} \! \sum_{t \in \mathcal{T}_{meta-test}} \sum_{i=1}^{N_t} \!-\!\log p_{\bm{\theta}^{*}}(\bm{y}_t^{(i)}|[\bm{I}_t; \bm{X}_t^{(i)}]) 
\end{split}
\label{eq:outer}
\end{align}
This optimization is called bilevel optimization and is commonly used in hyperparameter optimization. 
Note that we use the manually created instruction $\bm{I}_t$ to measure the meta-test performance because we aim to develop a model that can accept arbitrary human-created instructions. 

\subsection{Instruction Embedder}
\label{sec:instructionembedder}

This section presents a method for creating learnable instructions $\bm{I}_\phi$.
As shown in Figure \ref{fig:instructionoptimization} (left), the instruction embedder replaces manually created instructions with the embeddings of learnable instructions or prepends them to manually created instructions.
We consider the following two types of parameterizations of learnable instructions:

\paragraph{Direct Parameterization (DP)}

We parameterize the learnable instruction $\bm{I}_\phi$ by preparing a learnable matrix for each task: $\bm{I}_\phi = \bm{W}_t \in \mathcal{R}^{l \times d}$ where $l$ denotes the arbitrary length of a learnable instruction, and $d$ is the dimension of the embeddings in the model $\bm{\theta}$.
Although this parameterization is very simple, the size of the parameter $\bm{\phi}$ ($|\mathcal{T}_{train}| \times l \times d$) increases when many training tasks exist.
Moreover, as each learnable matrix $\bm{W}_t$ is updated only when task $t$ is used for computing the meta-train loss, the matrices are  updated infrequently when the number of training task is large.
Therefore, we propose another parameterization method that is scalable for a large number of training tasks. 

\paragraph{Instance Conversion (IC)}

Another parameterization method is to convert a task instance $\bm{z}_t^{(i)}$ into $\bm{I}_\phi$ as shown in Eq. \eqref{eq:encoder} and \eqref{eq:embed_ie}.
\begin{align}
\bm{h}_t^{(i)} &= \mathrm{avgpool}(\bm{z}_t^{(i)} \bm{V}_{\phi}) \label{eq:encoder} \\
\bm{I}_\phi &= \bm{W}_{\phi} \bm{h}_t^{(i)} \label{eq:embed_ie}
\end{align}
where the task instance $\bm{z}_t^{(i)}$ is a sequence of tokens defined as ``Input: $\bm{x}_t^{(i)}$ Output: $\bm{y}_t^{(i)}$'', where $\bm{x}_t^{(i)}$ and $\bm{y}_t^{(i)}$ represents the $i$-th input and output of a task $t$, respectively.
$\bm{V}_{\phi} \!\in\! \mathcal{R}^{v \times d'}$ is an word embedding matrix where $v$ denotes the vocabulary size, and $\mathrm{avgpool}$ denotes the average-pooling operation across the embedded tokens.
$\bm{h}_t^{(i)} \!\in\! \mathcal{R}^{d'}$ denotes a latent representation of $\bm{z}_t^{(i)}$, and $\bm{W}_{\phi} \!\in\! \mathcal{R}^{l \times d \times d'}$ is a learnable tensor to convert the latent representation into an instruction\footnote{We attempted to use T5 encoder for obtaining $\bm{h}_t^{(i)}$; however, it makes bilevel optimization unstable due to a large number of parameters.}.
We assume that $\bm{V}_{\phi}$ and $\bm{W}_{\phi}$ are optimized to generate an optimal instruction given a task instance.
As the parameters are shared across all training tasks, this parameterization is scalable for a large number of training tasks.

\subsection{Instruction Extractor}
\label{sec:instructionextractor}

We consider another type of instruxction that has multiple candidates to use. 
A task exemplar is one example because every task instance $j \!\in\! \{1, \ldots, N_t\}$ in the training set can be used as a task exemplar.
While instruction tuning randomly selects a task exemplar as instruction, an optimal task exemplar would exist for cross-task generalization.
We explore how to select the optimal task exemplar that maximizes the performance on unseen tasks.
An outline of the instruction extractor is shown in Figure \ref{fig:instructionoptimization} (right).

We parameterize the probability $p_{\phi}(\bm{z}_t^{(j)})$, where the $j$-th instance is selected as an exemplar of task $t$.
Similar to the instruction embedder, we consider the following two parameterizations:

\paragraph{Direct Parameterization (DP)}

We parameterize the logits of $p_{\phi}(\bm{z}_t^{(j)})$ by using a learnable vector $\bm{v}_t \in \mathcal{R}^{N_t}$ for each task $t$.
The logits are converted into probabilities using softmax function in Eq. \eqref{eq:prob_dp}.
\begin{align}
p_{\phi}(\bm{z}_t^{(j)}) = \frac{\exp(v_t^{(j)})}{\sum_{j=1}^{N_t} \exp(v_t^{(j)})}
\label{eq:prob_dp} 
\end{align}
This parameterization is simple but not scalable when the number of training tasks is large.

\paragraph{Instance Conversion (IC)}

While direct parameterization parameterizes $p_{\phi}(\bm{z}_t^{(j)})$ regardless of the task instance (i.e., task input and output), instance conversion considers the conditional probability given a task instance.
Specifically, instance conversion parameterizes the probability where $\bm{z}_t^{(j)}$ is selected as the exemplar of instance $\bm{z}_t^{(i)}$ in Eq. \eqref{eq:prob_ie}.
\begin{align}
p_{\phi}(\bm{z}_t^{(j)}|\bm{z}_t^{(i)}) = \frac{\exp(\bm{h}_t^{(j)}\bm{W}_{\phi}\bm{h}_t^{(i)})}{\sum_{j=1}^{N_t} \exp(\bm{h}_t^{(j)}\bm{W}_{\phi}\bm{h}_t^{(i)})} \label{eq:prob_ie} 
\end{align}
where $\bm{W}_{\phi} \!\in\! \mathcal{R}^{d' \times d'}$ denotes a learnable matrix, and $\bm{h}_t^{(j)} \!\in\! \mathcal{R}^{d'}$ is a latent representation of the task instance $\bm{z}_t^{(j)}$ obtained by Eq. \eqref{eq:encoder}.
This parameterization assumes that $\bm{V}_{\phi}$ and $\bm{W}_{\phi}$ are optimized to select an optimal exemplar given a task instance.
As the parameters $\bm{\phi}$ are shared across all training tasks, this parameterization is also scalable for a large number of training tasks.

\vspace{\baselineskip}

Subsequently, an instance with the highest probability is extracted as an instruction as shown in Eq. \eqref{eq:argmax} and \eqref{eq:embed}. 
\begin{align}
\bm{z}_t &= \argmax_j p_{\phi}(\bm{z}_t^{(j)}) \label{eq:argmax} \\
\bm{I}_{\phi} &= \bm{z}_t \bm{V}_{\theta} \label{eq:embed}
\end{align}
where $\bm{V}_{\theta} \!\in\! \mathcal{R}^{v \times d}$ is the word embedding matrix of the model $\bm{\theta}$.
Since $\argmax$ operation is not differentiable, we use the straight-through estimator \cite{bengio2013estimating} to approximate the gradient in the backward pass\footnote{We also tried to compute $\bm{I}_{\phi}$ using the expectation of $\bm{z}_t^{(j)}$: $\bm{I}_{\phi} \!=\! \mathbf{E}_{p_{\phi}}[\bm{z}_t^{(j)} \bm{V}_{\theta}]$ instead of $\argmax$ operation; however, it significantly underperforms.}.
As computing the probability of all instances requires a high computational cost when the number of instances is significant, we set a constant value as $N_t\!=\!N$ and randomly sampled $N$ instances from all training instances.

\begin{algorithm}[H]
    \begin{algorithmic}
    \WHILE{not converged}
    \FOR{$k = 1, \ldots, K$}
    \STATE $\bm{\theta}^{(k)} \leftarrow \bm{\theta}^{(k-1)} - \eta \substitute{\nabla_{\bm{\theta}} \mathcal{L}_{in}(\bm{\theta},\bm{\phi})}{\bm{\theta}=\bm{\theta}^{(k-1)}}$
    \ENDFOR
    \STATE $\bm{\phi} \leftarrow \bm{\phi} - \eta \nabla_{\bm{\phi}} \mathcal{L}_{out}(\bm{\theta}^{(K)})$
    \ENDWHILE
    \end{algorithmic}
    \caption{Bilevel Optimization}
    \label{alg}
\end{algorithm}

\subsection{Efficiently Solving Bilevel Optimization}

Directly solving bilevel optimization requires a substantial computational cost because it includes a nested formulation.
As shown in Alg. \ref{alg}, approximating the inner optimization in Eq. \eqref{eq:inner} by $K$-gradient steps significantly reduces the computational cost, where $K$ is large enough to reach the optimal points of the inner-loop \cite{franceschi2017forward, shaban2019truncated}.

Computing the hypergradient $\nabla_{\bm{\phi}} \mathcal{L}_{out}(\bm{\theta}^{(K)})$ still requires large memory space $\mathcal{O}(K|\bm{\theta}| \!+\! |\bm{\phi}|)$ as it needs to store $K$-step gradients \cite{franceschi2017forward}, and the language model $\bm{\theta}$ contains a lot of parameters. 
Using the implicit function theorem in Eq. \eqref{eq:hypergrad} and \eqref{eq:hessian}, the hypergradient can be computed without storing the intermediate gradients \cite{bengio2000gradient, lorraine2020optimizing}.

\vspace{-\baselineskip}
\begin{align}
&\nabla_{\bm{\phi}} \mathcal{L}_{out}(\bm{\theta}^{(K)}(\bm{\phi})) \!=\! \frac{\partial \mathcal{L}_{out}(\bm{\theta}^{(K)})}{\partial \bm{\theta}^{(K)}} \frac{\partial \bm{\theta}^{(K)}(\bm{\phi})}{\partial \bm{\phi}} \label{eq:hypergrad} \\
&\frac{\partial \bm{\theta}^{(K)}(\bm{\phi})}{\partial \bm{\phi}} \!=\! \substitute{- \Bigl[ \frac{\partial \mathcal{L}_{in}(\bm{\theta}, \bm{\phi})}{\partial \bm{\theta} \partial \bm{\theta}} \Bigr]^{\!-\!1} \frac{\partial \mathcal{L}_{in}(\bm{\theta}, \bm{\phi})}{\partial \bm{\theta} \partial \bm{\phi}}}{\bm{\theta}^{(K)},\bm{\phi}} \label{eq:hessian}
\end{align}

However, it is impractical to compute the inverse of the Hessian matrix in Eq. \eqref{eq:hessian} as exactly inverting Hessian often requires $\mathcal{O}(|\bm{\theta}|^3)$ computational cost.
We thus approximate the inverse-Hessian using the Neumann approximation, which is introduced in the hyperparameter optimization \cite{lorraine2020optimizing, zhang2021idarts}.
The inverse of the Hessian matrix can be approximated as shown in Eq. \eqref{eq:neumann}.

\vspace{-\baselineskip}
{\small
\begin{align}
&\Bigl[ \frac{\partial \mathcal{L}_{in}(\bm{\theta}, \bm{\phi})}{\partial \bm{\theta} \partial \bm{\theta}} \Bigr]^{\!-\!1} \!=\! \lim_{M \to \infty} \! \gamma \! \sum_{m=0}^{M} \! \Bigl[\bm{E} \!-\! \gamma \frac{\partial \mathcal{L}_{in}(\bm{\theta}, \bm{\phi})}{\partial \bm{\theta} \partial \bm{\theta}} \Bigr]^{m}
\label{eq:neumann}
\end{align}
}
where $\bm{E}$ denotes an identity matrix.
$\gamma \in \mathcal{R}$ is sufficiently small to satisfy $\|\bm{E} \!-\! \gamma \frac{\partial \mathcal{L}_{in}(\bm{\theta}, \bm{\phi})}{\partial \bm{\theta} \partial \bm{\theta}}\| < 1$ in the operator norm.
Consequently, the computational cost of the hypergradient considerably decreases to $\mathcal{O}(|\bm{\theta}| \!+\! |\bm{\phi}|)$ as shown in \citet{lorraine2020optimizing}.

\begin{figure*}[t!]
\centering
\includegraphics[width=\linewidth]{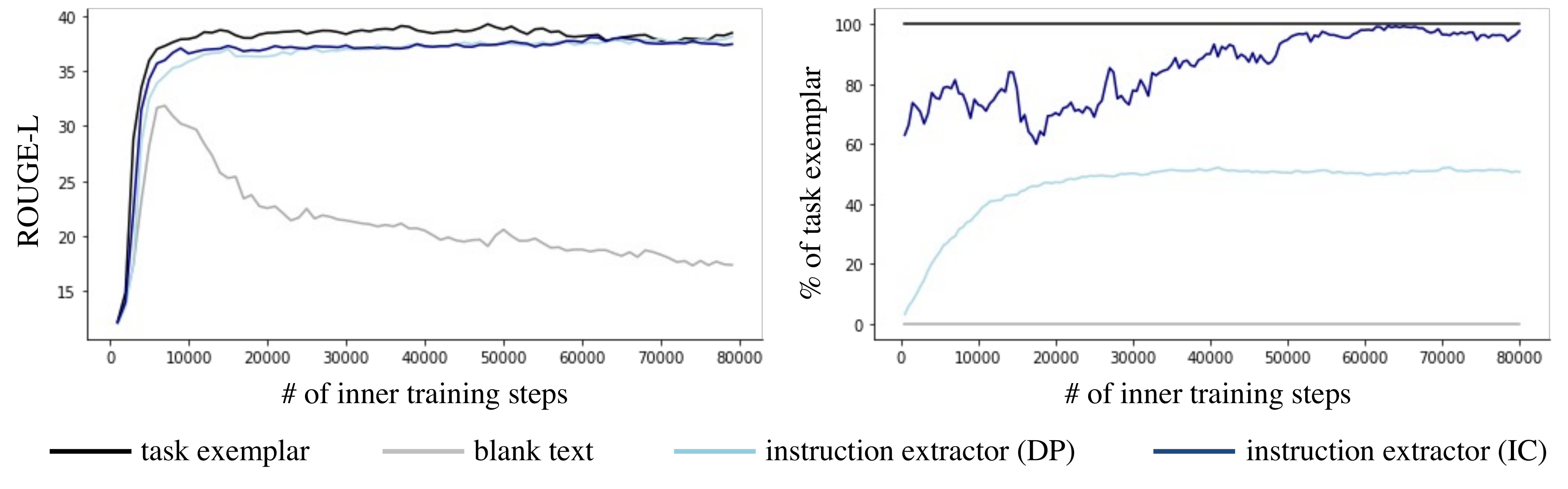}
\caption{Left: ROUGE-L on test tasks where a task exemplar is used as \emph{testing} instruction, while \emph{training} instruction is varied as above. Right: the percentage of training instances where a task exemplar is used as training instruction.}
\label{fig:blank}
\end{figure*}

\begin{table}[t!]
\small
\centering
\begin{tabularx}{\columnwidth}{p{2.3cm}RRRR}
\toprule
Split&Meta-train&Meta-test&Valid&Test\\ 
\midrule
\# of tasks&715&42&757&119 \\ 
\# of task types&50&10&60&12 \\ 
\# of instances/task&100&100&10&100 \\ 
\bottomrule
\end{tabularx}
\caption{Statistics of the dataset.}
\label{tbl:dataset}
\end{table}

\section{Experiments}

\subsection{Experimental Setup\footnote{The code is available at \url{https://github.com/misonuma/instopt}.}}

\paragraph{Dataset}
In this experiment, we used \textsc{Super-NaturalInstructions} \cite[\textsc{Sup-NatInst};][]{wang-etal-2022-super} as a benchmark to measure cross-task generalization.
\textsc{Sup-NatInst} consists of over 1,600 diverse tasks and their instructions across multiple languages.
We used English tasks and their instructions, resulting in 876 tasks in total.

We used the same test split of tasks (12 types; 119 tasks) and 100 instances for each task as \citet{wang-etal-2022-super}.
The remaining 60 task types (757 tasks) were used for meta-train, meta-test, and validation.
The validation set consisted of 10 instances across all 757 tasks, which were used to determine hyperparameters including meta-train/test split.
Based on the validation performance, we split the 60 task types into 50 and 10 types, which were used for the meta-train and meta-test set, respectively.
We used $100$ instances of each task for the meta-train/test set.
Table \ref{tbl:dataset} summarizes the statistics for each split.
The task types in each split are listed in Appendix \ref{app:tasksplit}.

\paragraph{Evaluation \& Baselines}
We assessed the cross-task generalization in two settings: a zero-shot setting that uses task definition as \emph{testing} instruction, and a one-shot setting that uses a task exemplar (n=1) as \emph{testing} instruction.
We adopted ROUGE-L \cite{lin-2004-rouge} to evaluate all tasks.
\citet{wang-etal-2022-super} shows that the human evaluation results align quite well with ROUGE-L across a variety of tasks.

For baseline training instructions, we used manually created instructions (e.g., task definition), exemplars randomly selected for each task or each instance. 
Learnable instructions induced by the instruction embedder or optimal exemplars selected by the instruction extractor were compared.

\paragraph{Implementation Details}

In our experiment, we used pretrained T5 \cite{raffel2020exploring} as the model $\bm{\theta}$.
Specifically, we use the LM-adapted version of the original T5-base (220M)\footnote{\url{https://huggingface.co/google/t5-base-lm-adapt}}, which is further trained with a language modeling objective \cite{lester-etal-2021-power}.
The hyperparameters of model $\bm{\theta}$ were tuned based on the validation performance of instruction tuning (baselines), and the same hyperparameters were used for instruction optimization.
The hyperparemters of learnable instructions $\bm{\phi}$ were determined w.r.t. the validation performance of instruction optimization.
Further details are provided in Appendix \ref{app:implementation}.

\subsection{Proof of Concept}

Before moving on to the comparison with instruction tuning, we show that our instruction extractor successfully optimizes the training instruction.
We trained models with two types of \emph{training} instructions: one of which is a task exemplar, and the other is a blank text.
Then, we evaluated them on the test set, where a task exemplar is used as the \emph{testing} instruction.
As shown in Figure \ref{fig:blank} (left), the model trained with a task exemplar achieves nearly 40\% ROUGE-L (black), whereas the model trained with blank text significantly declines to approximately 20\% ROUGE-L (gray).

Following these preliminary results, we verified that our instruction extractor appropriately selects a task exemplar from the two training instructions and obtains sufficient generalization ability.
Figure \ref{fig:blank} (left) shows that our instruction extractor achieves competitive performance with the model trained with a task exemplar.
Specifically, the instance conversion (IC; blue) converges faster than the direct parameterization (DP; light blue). 
Figure \ref{fig:blank} (right) presents the percentage of training instances where a task exemplar is selected as the training instruction.
Regarding the DP, the percentage increases smoothly, whereas it saturates at approximately 50\%.
In contrast, the IC reaches almost 100\%, though the increase is slightly unstable.
These results indicate that our instruction extractor successfully selects an appropriate training instruction.
Note that the training time of instruction optimization is reasonable compared to instruction tuning, as shown in Appendix \ref{app:computationaltime}.

\begin{table}[t!]
\small
\centering
\begin{tabularx}{\linewidth}{p{5.15cm}R}
\toprule
Training Instruction&ROUGE-L \\ 
\midrule
Def.                             &  33.82 $\pm$ 0.47 \\
Def. + Pos.                      &  27.74 $\pm$ 0.41 \\
Def. + Pos. + Neg.               &  27.91 $\pm$ 0.66 \\
Def. + Pos. + Neg. + Expl.       &  29.07 $\pm$ 0.31 \\
\midrule
Instruction Embedder (DP)        &  11.79 $\pm$ 0.27 \\
Instruction Embedder (IC)        &  11.99 $\pm$ 0.22 \\
Def. + Instruction Embedder (DP) &  34.79 $\pm$ 0.33 \\
Def. + Instruction Embedder (IC) &  \textbf{34.97 $\pm$ 0.46} \\
\bottomrule
\end{tabularx}
\caption{Zero-shot evaluation where task definition is used as \emph{testing} instruction, while \emph{training} instruction is varied as above. Def.: task definition; Pos.: positive exemplar (n=1), Neg.: negative exemplar (n=1); Expl.: explanation why each positive/negative exemplar is correct/incorrect. DP and IC represents direct parameterization and instance conversion, respectively. }
\label{tbl:result_definition}
\end{table}

\subsection{Main Results}
\label{sec:results}
Here, we examine the effectiveness of instruction optimization by comparing it with the baselines.
In Table \ref{tbl:result_definition} and \ref{tbl:result_exemplar}, we show the average performance across 8 different random seeds and 95\% confidence intervals w.r.t. the t-distribution.

Table \ref{tbl:result_definition} shows the average ROUGE-L across all test tasks where the task definition is used as the testing instruction, while varying the training instruction.
As the baseline of training instructions, we used manually created task definitions concatenated with positive/negative exemplars and explanations about each positive/negative exemplar.
When using only learnable instructions generated by the instruction embedder, the performance is considerably worse than that of baselines.
This underperformance suggests that the learned instructions cannot alternate with manually created instructions.
However, concatenating learnable instruction with task definition leads to performance gain, whereas prepending other instructions (positive/negative exemplars and explanations) has a negative effect.
As will be elaborated in Section \ref{sec:analysisinstruction}, adding learnable instructions improves the diversity of instructions and achieves higher generalization performance.

In Table \ref{tbl:result_exemplar}, we show the results where a task exemplar is used as the testing instruction.
Unfortunately, our instruction extractor underperforms exemplars randomly selected for each \emph{task} (i.e., the same exemplar is used for each instance).
To investigate the reason for the worse performance, we added another baseline, which randomly selects an exemplar for each \emph{instance} (i.e., different exemplars are used for each instance).
Unexpectedly, the random exemplars yield considerably worse ROUGE-L when they are selected for each instance.
This result indicates that using the same exemplar across all instances of each task is preferable for cross-task generalization.
As the instruction extractor (DP and IC) updates the optimal exemplar during the optimization, it performs worse than exemplars randomly selected for each task.
In particular, as IC varies the optimal exemplar for each instance, it results in a lower performance.

The evaluation results of each test task type are shown in Appendix \ref{app:result_task}.

\begin{table}[t!]
\small
\centering
\begin{tabularx}{\linewidth}{p{5.15cm}R}
\toprule
Training Instruction&ROUGE-L \\ 
\midrule
Random Exemplar (each task) &  \textbf{39.59 ± 0.14} \\
Random Exemplar (each instance) &  37.19 ± 0.25 \\
\midrule
Instruction Extractor (DP) &  37.85 ± 0.67 \\
Instruction Extractor (IC) &  37.15 ± 0.52 \\
\bottomrule
\end{tabularx}
\caption{One-shot evaluation where a task exemplar is used as \emph{testing} instruction while \emph{training} instruction is varied as above. Random Exemplar denotes exemplars randomly selected for each \emph{task} or each \emph{instance} (n=1). DP and IC represents direct parameterization and instance conversion, respectively. }
\label{tbl:result_exemplar}
\end{table}

\begin{figure*}[t!]
\centering
\includegraphics[width=\linewidth]{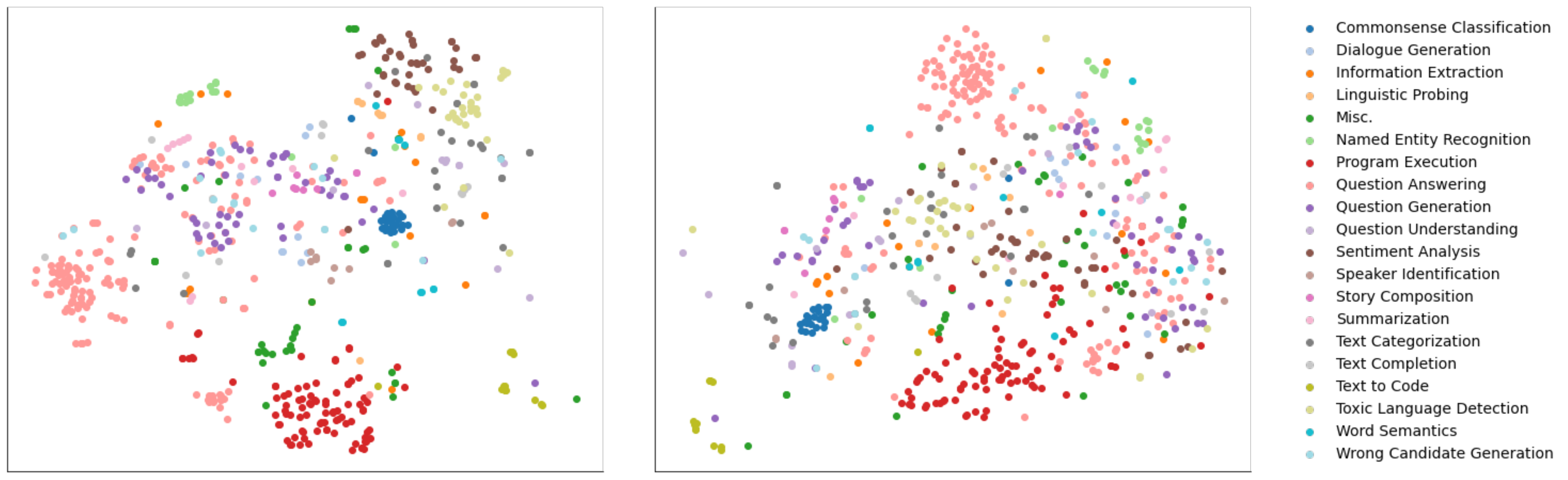}
\caption{Embeddings of the instructions in the meta-train set. Left: task definition; Right: learned instruction concatenated with task definition. Each point represents a task, while each color denotes the task type.}
\label{fig:instructionoembedding}
\end{figure*}

\section{Discussion}

\subsection{Analysis of Learned Instruction}
\label{sec:analysisinstruction}

We discuss how the learned instruction contributes to the improvement of cross-task generalization.

As the instruction embedder directly generates instruction embeddings in a continuous space, the learned instruction is difficult to interpret.
Following \citet{lester-etal-2021-power}, we computed the nearest neighbors of each token in the learned instruction from the vocabulary of the model $\bm{\theta}$; however, we could not find explicit patterns for the nearest tokens.
Therefore, we computed the embeddings of the learned instructions and visuzalized them at a two-dimensional space using t-SNE \cite{van2008visualizing}.
The embeddings were obtained by the average pooling across the last hidden states encoded by the T5 encoder.

In Figure \ref{fig:instructionoembedding}, we show the embeddings of top 20 task types with respect to the number of tasks in the meta-train set. 
The embeddings of the task definition (left) are closely clustered by the task type, and training tasks do not cover some spaces.
On the other hand, the embeddings of learned instructions (right) are roughly clustered, and some task types are scattered over the embedding space (e.g., sentiment analysis and toxic language detection).
As learned instructions enhance the diversity of instructions and cover a broader embedding space, the trained model can generalize to wider variety of instructions.
Thus, learned instructions improve the generalization performance on unseen tasks. 

\begin{figure}[t!]
\centering
\includegraphics[width=\linewidth]{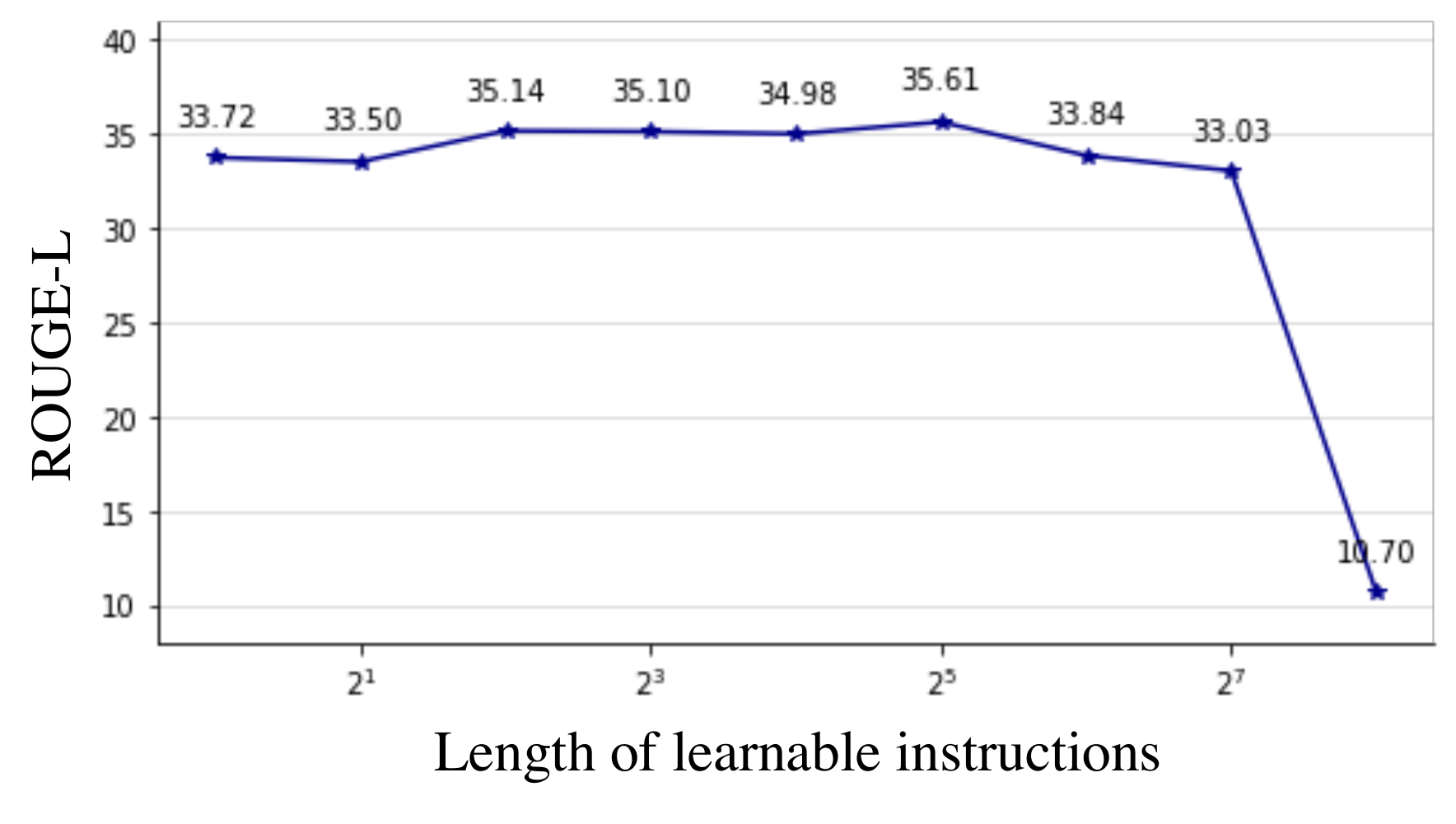}
\caption{ROUGE-L on the test set where the length of learnable instructions is varied.}
\label{fig:instruction_length}
\end{figure}

Figure \ref{fig:instruction_length} shows the generalization performance concerning the length of the learnable instruction prepended to the task definition.
The model’s performance saturates when the length is $2^6\!=\!64$. 
When the instruction is longer than $64$, the performance declines significantly.
As bilevel optimization tends to be unstable for large-scale hyperparameters, a large instruction length leads to low generalization performance.

\subsection{Analysis of Meta-train/test Split}

We study how meta-train/test split affects the generalization performance of the trained model.

\begin{figure}[t!]
\centering
\includegraphics[width=\linewidth]{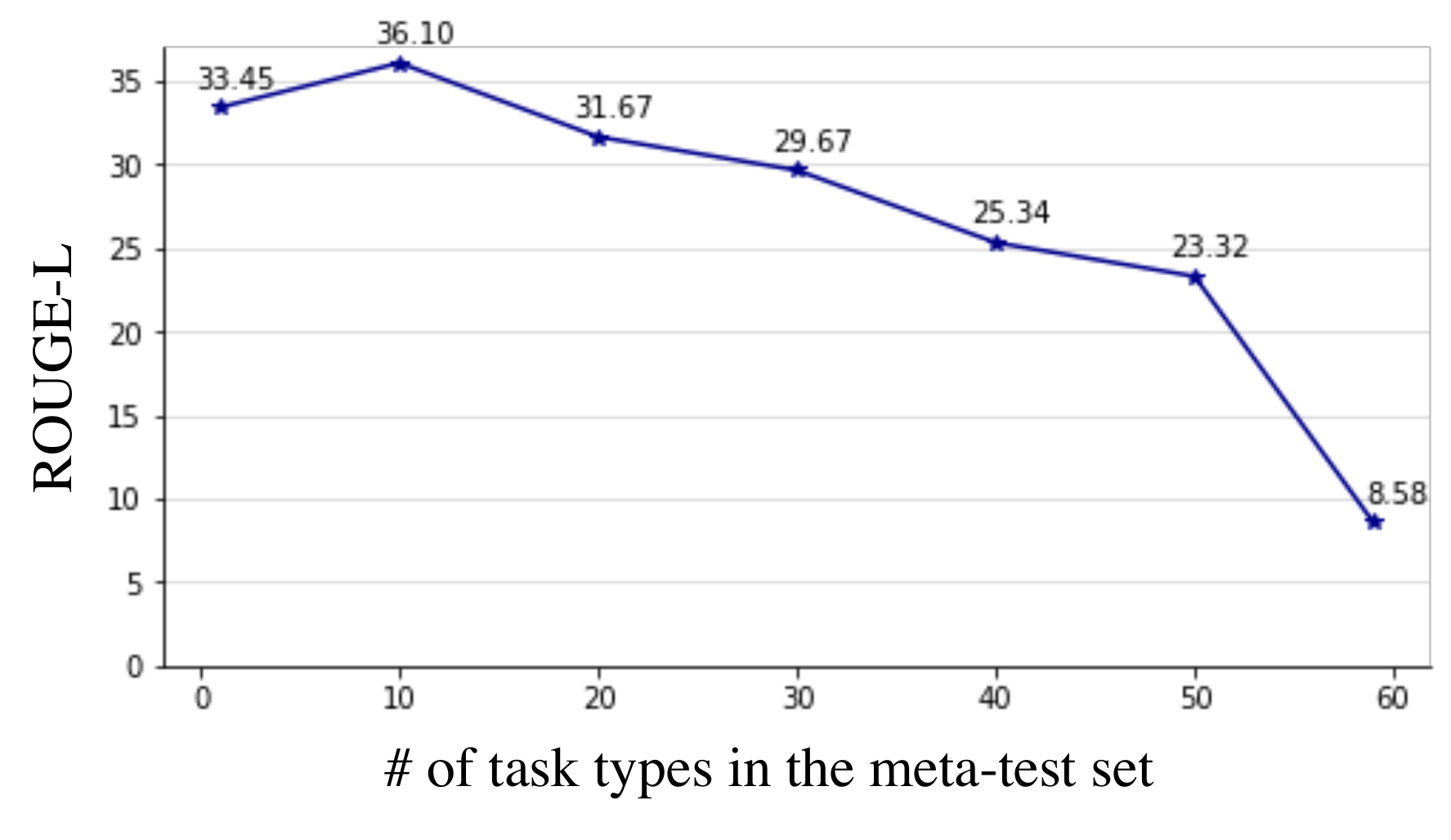}
\caption{ROUGE-L on the test set w.r.t. the number of task types in the meta-test set.}
\label{fig:n_category_meta}
\end{figure}

\paragraph{Number of Meta-train/test Tasks}
Figure \ref{fig:n_category_meta} shows the performance with different numbers of task types in the meta-train/test split: 1/59, 10/50, 20/40, 30/30, 40/20, 50/10, and 59/1.
In each split, meta-train/test tasks were randomly chosen.
The trained model achieves the best generalization performance when the number of categories in the meta-test is 10.
The performance worsens as the number of meta-test tasks increases, while the number of meta-train tasks decreases correspondingly.

\paragraph{Diverse vs. Not Diverse}

We examine whether meta-test tasks should be diverse or not diverse.
If meta-test tasks are diverse, the generalization performance would be improved because the instruction is trained to achieve higher performance on various tasks.
However, it also increases the risk that some of meta-test tasks are similar to meta-train tasks, which would negatively affect the performance on unseen tasks.
It is not obvious whether meta-test tasks should be diverse or not diverse. 

To answer this question, we prepared two types of meta-test splits.
One comprises randomly selected tasks, whereas the other consists of tasks that are grouped by k-means clustering.
We prepared 16 different random splits, while k-means divided the tasks into 16 groups based on the embeddings of the task definition.
Then, for both random split and k-means, the best split for the validation set was chosen from the 16 splits.
Experimental results show that model trained on the random split achieves 36.1 ROUGE-L, while that of k-means scores 35.0 ROUGE-L on the test set.
Although the margin is not significant, we confirmed that diverse meta-test tasks are more preferable for cross-task generalization.

\section{Related Work}

\paragraph{Instruction Tuning}

Instruction tuning has attracted considerable attention to achieve models that are generalizable across a variety of tasks \cite{wei2021finetuned,sanh2021multitask,mishra-etal-2022-cross}.
By prepending either a few exemplars \cite{min-etal-2022-metaicl, chen-etal-2022-meta} or text-based instructions \cite{wei2021finetuned,sanh2021multitask,mishra-etal-2022-cross} to multi-task learning, the trained model can generalize to tasks unseen during training.
Further progress has been made by scaling the number of tasks \cite{wang-etal-2022-super, chung2022scaling}, scaling the model size \cite{chung2022scaling, scao2022bloom}, and improving the training strategy \cite{lang2022co, min-etal-2022-noisy, ye2022guess}.
In contrast, our work is the first study to optimize training instructions to improve the cross-task generalization ability.

Although \textsc{Super-NaturalInstructions} \cite{wang-etal-2022-super} is used as the benchmark for measuring cross-task generalization in our study, our instruction optimization can be applied to other cross-task benchmarks, such as CROSSFIT \cite{ye-etal-2021-crossfit} and PromptSource \cite{bach-etal-2022-promptsource}.

\paragraph{Prompt Engineering}

Recent instruction-based NLP has evolved prompt engineering, which seeks the most appropriate prompt to achieve a task \cite{liu2022pretrain}.
While there are numerous studies to search for an optimal prompt in a discrete token space \cite{shin-etal-2020-autoprompt, schick-schutze-2021-exploiting, gao-etal-2021-making}, some work studies continuous prompts that perform prompting in the embedding space of tokens \cite{li-liang-2021-prefix, lester-etal-2021-power, qin-eisner-2021-learning}.
Other studies retrieve appropriate exemplars as a testing prompt for in-context learning and achieve better performance than randomly selected exemplars \cite{das-etal-2021-case, liu-etal-2022-makes, rubin-etal-2022-learning}.
Whereas the aforementioned methods optimize prompts to achieve an individual task in the test, our study differs in the target and aim of optimization; we optimize the training prompts to maximize the generalization performance of the trained model.

\paragraph{Bilevel Optimization}

Bilevel optimization has been used to optimize hyperparameters \cite{franceschi2017forward, lorraine2020optimizing}, initial model weights \cite{finn2017model, franceschi2018bilevel}, and model architectures \cite{liu2018darts, zhang2021idarts}.
We optimize the training instructions by regarding them as a special type of hyperparameters.
Learnable instructions are constructed by many hyperparameters, which makes bilevel optimization difficult in terms of computational cost and stability.
Recent studies \cite{rajeswaran2019meta, lorraine2020optimizing, zhang2021idarts} significantly reduce the computational cost and improve the stability by combining the implicit function theorem with efficient inverse Hessian approximations.
We leverage this idea for instruction optimization, achieving instruction optimization at a reasonable computational cost and stability.

\section{Conclusion}

This study presents instruction optimization, which optimizes training instructions concerning generalization ability.
The experimental results showed that our instruction extractor successfully extracted appropriate instruction, providing proof of concept.
Regarding the comparison with instruction tuning, the instruction embedder enhanced the diversity of instructions and improved the generalization ability than using only manually created instructions.
In contrast, the instruction extractor did not contribute to the performance gain because using the same task exemplar across instances is unexpectedly preferable for cross-task generalization.
This study provides a basis for exploring the optimal instructions for instruction tuning.

\section*{Limitations}

Our study used T5-base (220M) due to the capacity of our computational resources (Tesla V100 32GB).
Thus, it is unclear whether our method is also effective for larger models, such as T5-XL/XXL.
\citet{lester-etal-2021-power} argues that continuous prompts are particularly effective for large T5 models.
Following their results, our instruction embedder is also expected to be effective for larger models.

As shown in Figure \ref{fig:blank}, instruction optimization is slightly unstable to converge.
Some studies tackled the unstable convergence of bilevel optimization by L2-normalization, early stopping \cite{zela2019understanding}, or perturbation of hyperparameters \cite{chen2020stabilizing}.
These methods might be effective in stabilizing the instruction optimization.

\section*{Ethics Statement}

Our study complies with the ACL Ethics Policy.
We used S2ORC \cite[][CC BY-NC 4.0]{lo-etal-2020-s2orc}, PyTorch \cite[][BSD-style license]{NEURIPS2019_9015} and HuggingFace Transformers \cite[][Apache-2.0]{wolf-etal-2020-transformers} as scientific artifacts.
Our study was conducted under the licenses and terms of the scientific artifacts.
Our model is trained on a set of publicly available datasets \cite{wang-etal-2022-super}, in which undesirable data distribution, such as disinformation, bias, or offensive content, might present.
Such potential risks need to be recognized.

\section*{Acknowledgements}

We would like to thank the anonymous reviewers for their valuable feedback.
This work was supported by JST ACT-X JPMJAX1904, JST CREST JPMJCR21D1, NEDO JPNP20006, and JSPS KAKENHI 23K16940, Japan.

\bibliography{anthology,custom}
\bibliographystyle{acl_natbib}

\clearpage
\appendix

\section{Appendix}

\subsection{Task Split}
\label{app:tasksplit}

The task types used in the meta-train/meta-test/test split are listed in Table \ref{tbl:tasks}.
We prepared 16 random splits of meta-train/test and used the one that achieved the best validation performance.

\begin{table}[t!]
\small
\centering
\scalebox{0.86}{
\begin{tabularx}{\linewidth}{p{5.15cm}R}
\toprule
Task types in meta-train set & \# of tasks \\ 
\midrule
Answer Verification                 &         3 \\
Code to Text                        &         4 \\
Coherence Classification            &         6 \\
Commonsense Classification          &        23 \\
Dialogue Generation                 &        11 \\
Dialogue State Tracking             &         4 \\
Discourse Connective Identification &         1 \\
Entity Generation                   &         1 \\
Fill in The Blank                   &         8 \\
Gender Classification               &         7 \\
Grammar Error Detection             &         2 \\
Information Extraction              &        17 \\
Irony Detection                     &         2 \\
Linguistic Probing                  &         9 \\
Mathematics                         &         4 \\
Misc.                               &        36 \\
Named Entity Recognition            &        17 \\
Negotiation Strategy Detection      &         7 \\
Number Conversion                   &         2 \\
Paraphrasing                        &         4 \\
Poem Generation                     &         1 \\
Pos Tagging                         &         9 \\
Program Execution                   &        90 \\
Punctuation Error Detection         &         1 \\
Question Answering                  &       158 \\
Question Decomposition              &         2 \\
Question Generation                 &        51 \\
Question Understanding              &        13 \\
Sentence Composition                &         7 \\
Sentence Compression                &         1 \\
Sentence Expansion                  &         1 \\
Sentence Ordering                   &         3 \\
Sentence Perturbation               &         4 \\
Sentiment Analysis                  &        42 \\
Spam Classification                 &         1 \\
Speaker Identification              &         9 \\
Speaker Relation Classification     &         2 \\
Story Composition                   &         9 \\
Style Transfer                      &         2 \\
Summarization                       &        12 \\
Text Categorization                 &        28 \\
Text Completion                     &        14 \\
Text Quality Evaluation             &         7 \\
Text Simplification                 &         4 \\
Text to Code                        &        12 \\
Toxic Language Detection            &        32 \\
Translation                         &         2 \\
Word Relation Classification        &         5 \\
Word Semantics                      &        10 \\
Wrong Candidate Generation          &        15 \\
\midrule
Task types in meta-test set &  \# of tasks \\
\midrule
Discourse Relation Classification &         1 \\
Entity Relation Classification    &         1 \\
Explanation                       &         5 \\
Fact Verification                 &         3 \\
Intent Identification             &         4 \\
Preposition Prediction            &         1 \\
Spelling Error Detection          &         1 \\
Stance Detection                  &         2 \\
Stereotype Detection              &         7 \\
Text Matching                     &        17 \\
\midrule
Task types in test set &  \# of tasks \\
\midrule
Answerability Classification &        13 \\
Cause Effect Classification  &         7 \\
Coreference Resolution       &        14 \\
Data to Text                 &         9 \\
Dialogue Act Recognition     &         7 \\
Grammar Error Correction     &         1 \\
Keyword Tagging              &         5 \\
Overlap Extraction           &         2 \\
Question Rewriting           &        11 \\
Textual Entailment           &        24 \\
Title Generation             &        18 \\
Word Analogy                 &         8 \\
\bottomrule
\end{tabularx}
}
\caption{Task types used in each split.}
\label{tbl:tasks}
\end{table}

\begin{table*}[t!]
\small
\begin{tabularx}{\textwidth}{p{4.0cm}R|RRRR} 
\toprule
Training Instruction& Def. & Inst. Emb. (DP) & Inst. Emb. (IC) & Def. + Inst. Emb. (DP) & Def. + Inst. Emb. (IC) \\
\midrule
Answerability Classification &  41.20 ± 0.66 &               8.67 ± 0.79 &               9.84 ± 0.52 &                     41.21 ± 0.47 &                     41.13 ± 0.56 \\
Cause Effect Classification  &  49.77 ± 0.42 &              15.80 ± 1.84 &              16.35 ± 2.03 &                     50.47 ± 0.62 &                     50.36 ± 0.74 \\
Coreference Resolution       &  32.30 ± 2.16 &              12.09 ± 0.75 &              11.14 ± 0.56 &                     34.03 ± 0.91 &                     33.79 ± 0.54 \\
Data To Text                 &  27.51 ± 0.49 &              13.61 ± 0.91 &              13.43 ± 0.70 &                     29.45 ± 0.46 &                     29.35 ± 0.55 \\
Dialogue Act Recognition     &  35.95 ± 3.76 &               8.23 ± 1.08 &               8.61 ± 0.98 &                     36.58 ± 2.63 &                     35.73 ± 4.05 \\
Grammar Error Correction     &  85.20 ± 0.28 &              79.27 ± 1.92 &              76.20 ± 3.48 &                     85.13 ± 0.21 &                     85.09 ± 0.08 \\
Keyword Tagging              &  49.52 ± 1.36 &              19.94 ± 1.71 &              19.69 ± 1.04 &                     50.62 ± 1.64 &                     50.96 ± 1.14 \\
Overlap Extraction           &  20.94 ± 0.41 &              18.13 ± 0.48 &              17.49 ± 1.25 &                     20.64 ± 0.45 &                     21.27 ± 0.69 \\
Question Rewriting           &  43.28 ± 1.52 &              14.95 ± 1.21 &              15.90 ± 0.78 &                     45.49 ± 1.72 &                     45.76 ± 1.98 \\
Textual Entailment           &  34.68 ± 2.21 &               7.46 ± 0.83 &               8.03 ± 0.55 &                     36.36 ± 0.83 &                     37.37 ± 0.94 \\
Title Generation             &  21.55 ± 0.29 &              13.02 ± 0.86 &              12.94 ± 0.35 &                     21.50 ± 0.36 &                     21.55 ± 0.30 \\
Word Analogy                 &  14.01 ± 1.21 &               4.88 ± 0.84 &               4.88 ± 0.63 &                     13.46 ± 1.00 &                     13.70 ± 0.31 \\
\midrule
Average                      &  33.82 ± 0.47 &              11.79 ± 0.27 &              11.99 ± 0.22 &                     34.79 ± 0.33 &                     34.97 ± 0.46 \\
\bottomrule
\end{tabularx}
\caption{Zero-shot evaluation where task definition is used as \emph{testing} instruction, while \emph{training} instruction is varied as above. Def.: task definition; Inst. Emb.: Instruction Embedder. DP and IC represents direct parameterization and instance conversion, respectively. }
\label{tbl:result_definition_each_task}
\end{table*}

\begin{table*}[t!]
\small
\begin{tabularx}{\textwidth}{p{4.0cm}RR|RR} 
\toprule
Training Instruction & Random Exemplar (each task) & Random Exemplar (each instance) & Instruction Extractor (DP) & Instruction Extractor (IC) \\
\midrule
Answerability Classification &     52.79 ± 0.43 &    53.27 ± 0.55 &               53.18 ± 0.59 &               53.24 ± 0.68 \\
Cause Effect Classification  &     53.22 ± 0.26 &    53.16 ± 0.37 &               52.63 ± 0.37 &               52.21 ± 0.49 \\
Coreference Resolution       &     41.59 ± 0.55 &    37.70 ± 0.62 &               37.27 ± 0.95 &               36.63 ± 0.54 \\
Data To Text                 &     37.29 ± 0.19 &    37.04 ± 0.22 &               37.31 ± 0.40 &               37.15 ± 0.20 \\
Dialogue Act Recognition     &     36.24 ± 0.43 &    33.56 ± 0.73 &               35.47 ± 1.06 &               36.33 ± 0.69 \\
Grammar Error Correction     &     85.35 ± 0.14 &    85.21 ± 0.06 &               85.13 ± 0.18 &               84.86 ± 0.26 \\
Keyword Tagging              &     52.96 ± 0.57 &    49.70 ± 1.47 &               50.63 ± 1.36 &               50.62 ± 2.18 \\
Overlap Extraction           &     33.45 ± 1.14 &    29.63 ± 2.17 &               32.64 ± 2.23 &               30.34 ± 1.55 \\
Question Rewriting           &     63.70 ± 0.59 &    64.66 ± 0.19 &               63.39 ± 1.31 &               63.24 ± 0.56 \\
Textual Entailment           &     31.70 ± 0.36 &    24.81 ± 1.05 &               27.07 ± 3.46 &               24.15 ± 1.89 \\
Title Generation             &     26.06 ± 0.27 &    24.25 ± 0.47 &               25.44 ± 0.31 &               25.29 ± 0.29 \\
Word Analogy                 &     16.11 ± 0.34 &    15.84 ± 0.56 &               16.03 ± 0.78 &               16.43 ± 0.33 \\
\midrule
Average                      &     39.59 ± 0.14 &    37.19 ± 0.25 &               37.85 ± 0.67 &               37.15 ± 0.52 \\
\bottomrule
\end{tabularx}
\caption{One-shot evaluation where a task exemplar is used as \emph{testing} instruction, while \emph{training} instruction is varied as above. Random Exemplar denotes exemplars randomly selected for each \emph{task} or each \emph{instance} (n=1).}
\label{tbl:result_exemplar_each_task}
\end{table*}

\subsection{Implementation Details}
\label{app:implementation}

We trained model $\bm{\theta}$ for three epochs using Adam \cite{kingma2014adam} with a learning rate of $1.0\!\times\!10^{-5}$ with linear decay, warmup steps of $8000$, and a batch size of $2$.
The maximum input and output length were set to $1024$ and $128$, respectively.

Learnable instructions $\bm{\phi}$ were trained using Adam with a batch size of $8$.
The learning rate was set to $1.0\!\times\!10^{-5}$ for instruction embedder (DP), $1.0\!\times\!10^{-6}$ for instruction embedder (IC), $5.0\!\times\!10^{-5}$ for instruction extractor (DP), $1.0\!\times\!10^{-5}$ for instruction extractor (IC) with linear decay.
The length of learnable instruction was $l\!=\!64$, the number of inner optimization steps was $K\!=\!20$ in Alg. \ref{alg}, the hyperparameters for the Neumann approximation were $M\!=\!1$ and $\gamma\!=\!1.0\!\times\!10^{-5}$ in Eq. \eqref{eq:neumann}.
The maximum input length in Eq. \eqref{eq:encoder} was $128$, and we randomly sampled $N\!=\!32$ instances for the candidates of the instruction extractor.

Our code is implemented with Python v3.8.13, PyTorch v1.12.0 \cite{NEURIPS2019_9015}, and transformers v4.18.0 \cite{wolf-etal-2020-transformers}.
Our code is based on the script published by \citet{wang-etal-2022-super}\footnote{\url{https://github.com/yizhongw/Tk-Instruct}}.
ROUGE-L is computed using the Python package distributed by Google\footnote{\url{https://pypi.org/project/rouge-score/}}.

\subsection{Computatinal Time}
\label{app:computationaltime}

Our experiments were conducted with a single Tesla V100 (32GB).
Each training run takes approximately 8 hours for instruction optimization, while it takes 5 hours for instruction tuning, without  validation.
However, the training time of instruction optimization depends on the number of inner training steps $K$.
It reduces to 6 hours when $K\!=\!100$, while slightly deteriorating the performance.

\subsection{Experimental Results for Each Test Task}
\label{app:result_task}

Table \ref{tbl:result_definition_each_task} and Table \ref{tbl:result_exemplar_each_task} shows the zero-shot and one-shot evaluation for each test task type, respectively.
We show the average performance across 8 different random seeds and 95\% confidence intervals w.r.t. the t-distribution.

\end{document}